\documentclass{llncs}
\usepackage[T1]{fontenc}
 \usepackage{booktabs} 
\usepackage{graphicx}
\usepackage{enumitem}
\usepackage{amsfonts}
\usepackage{hyperref}
\usepackage{xcolor}
\usepackage{float}

\usepackage[most]{tcolorbox}  
\usepackage{amsmath}          
\usepackage{amssymb}          

\begin{document}
\title{
Can LLMs extract human-like fine-grained evidence for  evidence-based fact-checking?
}
%
%
\author{
Antonín Jarolím\thanks{Correspondence to: \texttt{ijarolim@fit.vut.cz}}, 
Martin Fajčík and Lucia Makaiová 
}
%
%
\institute{Brno University of Technology, Czech Republic}

\maketitle              
\begin{abstract}

Misinformation frequently spreads in user comments under online news articles, highlighting the need for effective methods to detect factually incorrect information. To strongly support or refute claims extracted from such comments, it is necessary to identify relevant documents and pinpoint the exact text spans that justify or contradict each claim. This paper focuses on the latter task --- fine-grained evidence extraction for Czech and Slovak claims. We create new dataset, containing two-way annotated fine-grained evidence created by paid annotators. We evaluate large language models (LLMs) on this dataset to assess their alignment with human annotations. The results reveal that LLMs often fail to copy evidence verbatim from the source text, leading to invalid outputs. Error-rate analysis shows that the \texttt{llama3.1:8b} model achieves a high proportion of correct outputs despite its relatively small size, while the \texttt{gpt-oss-120b} model underperforms despite having many more parameters. Furthermore, the models \texttt{qwen3:14b}, \texttt{deepseek-r1:32b}, and \texttt{gpt-oss:20b} demonstrate an effective balance between model size and alignment with human annotations.

\end{abstract}
\section{Introduction}

On average, around three quarters of more than twelve thousand respondents across news sites expressed interest in having topic experts respond to comments in articles~\cite{stroud2017comment}.  
Unfortunately, this is highly impractical and therefore, an automatic discussion management is required. Major platforms have already recognized this need --- for example, platform \emph{X} (formerly Twitter) relies on human-generated “Community Notes”, while \emph{Seznam.cz} supplements selected comments with fact-checked contextual information.

Another promising direction for managing discussions is the automatic provision of relevant documents drawn from collections of trustworthy sources such as Reuters. 
However, locating the precise evidence within a relevant document can be time-consuming, and providing the entire document as evidence is typically too coarse to be persuasive. 
Therefore, identifying fine-grained evidence within the document is crucial. 
Readers can evaluate a short, well-targeted text span more quickly without decreasing the accuracy of judgement~\cite{ExtractiveExplanations-SelectAndRank}.
Consequently, the fine-grained evidence is both \emph{efficient} and \emph{effective} for mitigating misinformation.


To automate the management of online discussions, large language models (LLMs) can be employed to extract claims from user comments. A retrieval model can then identify relevant documents, after which LLMs can locate fine-grained evidence within the retrieved content.
This study focuses exclusively on the latter step --- identifying fine-grained evidence in source documents. Specifically, the contributions of this paper include:

\begin{enumerate}
    \item A \emph{manually constructed} dataset comprising check-worthy claims written in Czech or Slovak languages is constructed, each paired with a highly relevant document identified by annotators using various search engines. Two annotators then highlight fine-grained evidence supporting each claim.
    \item  Fine-grained evidence is identified using several LLMs, including the 685B \texttt{DeepSeek-R1} and 120B \texttt{gpt-oss} reasoning models, as well as a range of smaller open-weight models such as 27 billion (B) \texttt{Gemma-3} and 14 B \texttt{Phi4}.
    \item The performance of the LLMs is analysed in terms of error rates and alignment with human annotations in the context of fine-grained evidence generation.
\end{enumerate}

In this work, we focus on the analysis of \emph{supporting evidence}\footnote{The same kind of analysis can be performed on refuting evidence. We leave this to future work.} only.

\section{Related work}

\begin{description}[style=unboxed,leftmargin=0em,listparindent=\parindent]
\item [Automatic fact-checking.]
FactLens~\cite{mitra2025factlens} decomposes complex claims into sub-claims and evaluates the veracity of each independently.
Loki~\cite{li2025loki} extends this approach through an automated pipeline that identifies check-worthy claims, retrieves evidence, and verifies them.
However, in both methods, the evidence used to determine a claim’s truthfulness is provided only at the passage level.

Furthermore, AmbiFC~\cite{glockner2024ambifc} introduces ambiguity into automated fact-checking by incorporating multiple sentence-level annotations with potentially divergent labels.
It shows that models learning soft-label distributions for sentence-level evidence selection and veracity prediction achieve superior performance. This emphasizes the importance of fine-grained evidence for improving fact-checking accuracy.

\end{description}

\begin{description}[style=unboxed,leftmargin=0em,listparindent=\parindent]
    \setlength\parskip{0em}
\item [Fact-checking datasets.]

While AmbiFC~\cite{glockner2024ambifc} claims to provide fine-grained evidence alongside claims, its annotations remain at the passage level.
Other datasets, such as FEVER~\cite{thorne2018fever} and SciFact~\cite{wadden2020factorfictom-scifact}, offer finer granularity, as both include claims paired with sentence-level evidence.
The former focuses on general claims derived from Wikipedia, whereas the latter provides rationale annotations within scientific paper abstracts.

However, no comparable dataset exists for Czech or Slovak data, and, to our knowledge, none has evidence annotation at the span level. 
To address this gap, we construct a new dataset comprising Czech and Slovak samples with manually annotated fine-grained evidence spans.

\end{description}

\begin{description}[style=unboxed,leftmargin=0em,listparindent=\parindent]
    \setlength\parskip{0em}
\item [Reasoning capabilities of LLMs.]
The performance of LLMs continues to improve with increasing model size, architectural advancements~\cite{liu2025notjustscalinglaws}, and enhanced reasoning capabilities~\cite{guo2025deepseek}.
Recent work further demonstrates that LLMs exhibit strong reasoning performance on Czech and Slovak datasets~\cite{fajcik2025benczechmark}.

These developments open the possibility of employing LLMs for automatic fine-grained evidence extraction.
As a first step toward automated fact-checking with fine-grained evidence, we evaluate the alignment between human annotators and LLMs on this task.
\end{description} 

\section{Method}

This section outlines the methods for fine-grained evidence extraction. Given a claim and its associated text, the task is formally defined as follows:
\begin{description}[style=unboxed,leftmargin=0em,listparindent=\parindent]
    \setlength\parskip{0em}
\label{def: fine-grained evidence selection}
    \setlength\parskip{0em}
    \item[Problem statement.]
    Given a \emph{claim} and a tokenized \emph{text} $t = (t_1, t_2, \ldots, t_N)$, select a set of spans $S = \{ s_1, s_2, \ldots, s_M \}$, where $N, M \in \mathbb{N}^+$ and each span $s_m = (t_i, \ldots, t_j)$, with $i, j \in \mathbb{N}^+$ and $i <= j$, denotes a contiguous subsequence of $t$ that supports the claim.
\end{description}
The following section outlines how this task is approached by human annotators, baselines, and LLMs.

\begin{description}[style=unboxed,leftmargin=0em,listparindent=\parindent]
    \setlength\parskip{0em}
    \item[Two-Way Annotation of Fine-Grained Evidence Dataset.]

We collected 186 claim–text pairs and obtained two independent fine-grained evidence annotations for each sample, created by different annotators from a pool of eight non-expert annotators. 
The first annotation was produced using a custom annotation tool, while the second was created in Label Studio\footnote{\url{https://labelstud.io/}}. Annotation guidelines stated:

\begin{quote}
Highlight the \emph{minimal} portion of text that supports or refutes the claim.
Highlight the part that most convinces you that the given statement is \emph{true}.   
\end{quote}
Fortunately, both annotation interfaces allowed annotators to directly highlight evidence spans. Thus, each selected subsequence corresponds to a contiguous span appearing in the text. In contrast, LLMs were required to regenerate the selected span, allowing generation of subsequences not appearing in the text, as explained below.

\end{description}

\begin{description}[style=unboxed,leftmargin=0em,listparindent=\parindent]
    \setlength\parskip{0em}
    \item[Evidence Extraction using LLMs.]

We employ a diverse set of LLMs of varying sizes to perform fine-grained evidence extraction. Specifically, we use
\texttt{qwen2.5}~(72B, 32B)~\cite{qwen2025qwen25technicalreport},
\texttt{llama3.3}~(70B)~\cite{grattafiori2024llama},
\texttt{gemma2}~(27B)~\cite{team2024gemma},
\texttt{phi4}~(14B)~\cite{abdin2024phi},
\texttt{llama3.1}~(8B)~\cite{team2024gemma},
\texttt{gemma3}~(27B, 12B, 4B)~\cite{gemmateam2025gemma3technicalreport} and
\texttt{mixtral}~(8×7B)~\cite{jiang2024mixtralexperts}.
Additionally, we include Chain-of-Thought (CoT) LLMs that produce intermediate reasoning steps before providing the final answer
\texttt{gpt-oss-120b}~(20B, 120B)~\cite{openai2025gptoss120bgptoss20bmodel},
\texttt{deepseek-r1}~(685B, 32B)~\cite{guo2025deepseek},
\texttt{qwen3}~(32B, 14B)~\cite{yang2025qwen3}.

Similarly to the human annotators, the LLMs were presented with claim–text pairs, along with the comment from which the claim was originally extracted to provide additional context.
The models were instructed to identify the smallest possible segments of the text that directly support the claim and to output a JSON-formatted list of the selected spans (see Appendix~\ref{app:LLM Prompt} for the complete prompt). 
Although the LLMs were explicitly instructed to generate only spans appearing verbatim in the text, this constraint was not enforced technically. 
Consequently, as discussed in the following section, the models occasionally produced spans that did not occur in the source text.

\end{description}

\begin{description}[style=unboxed,leftmargin=0em,listparindent=\parindent]
    \setlength\parskip{0em}
    \item[Baseline Approaches.]
Additionally, we include non-neural approaches. Firstly, the \emph{claim baseline} tokenizes the claim $c$ into word tokens $c = (c_1, c_2, \ldots, c_O)$ and overlaps them with the corresponding text $t$, constructing the set of selected spans
$S_C = \{ (t_i, \ldots, t_j)| \exists k, l \in \mathbb{N}:  (t_i, \ldots, t_j) = (c_l, \ldots, c_k) \land l <= k \}$
.
The queries used to search for claim evidences during the fine-grained evidence annotations were also stored. 
These queries form the basis of the \emph{query baseline}, which overlaps the text with the query in the same manner as the claim baseline. 
Finally, the \emph{random baseline} uniformly samples contiguous spans whose number and length match those of a randomly chosen annotator for each sample.
\end{description}

\section{Results}

All experiments are conducted on the manually annotated dataset described above, comprising 186 samples with two independent sets of fine-grained evidence annotations.

\begin{description}[style=unboxed,leftmargin=0em,listparindent=\parindent]
    \setlength\parskip{0em}
    \item[Error Analysis of LLM-Generated Evidence.]
    
\label{sec:err_rate}

\begin{figure}
    \centering
    \includegraphics[width=\linewidth]{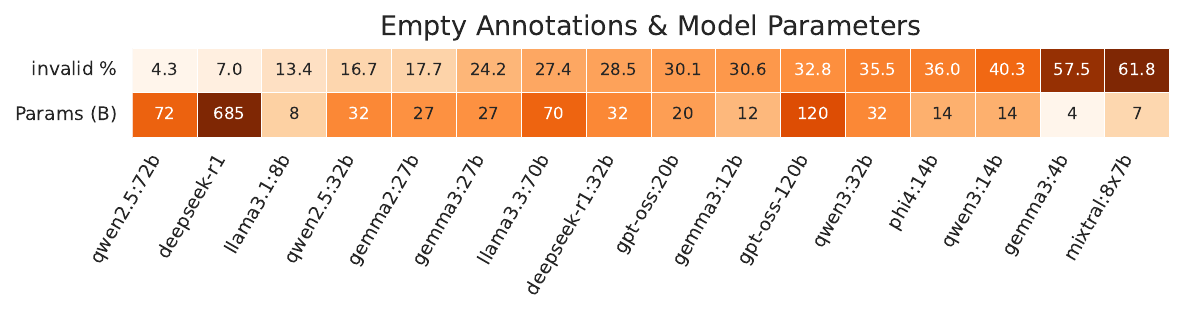}
    \caption{Various LLMs used to generate fine-grained extraction, their number of parameters in billions, and the percentage of incorrectly annotated samples (invalid \%) they generated.}
    \label{fig:err-rates}
\end{figure}

In general, model performance tends to improve with the number of parameters \cite{grattafiori2024llama,liu2025notjustscalinglaws}.
This trend is also evident in the present task, where the inability to generate valid output is interpreted as a failure to follow instructions.

The error rates of all evaluated LLMs are presented in Figure~\ref{fig:err-rates}. 
The highest number of incorrect annotations 61.8\% was produced by the \texttt{mixtral:8x7b} model. This result is consistent with the fact that \texttt{mixtral:8x7b} is among the smallest models in the comparison. Conversely, the \texttt{qwen2.5:72b} model achieved the lowest error rate, which aligns with its considerably larger parameter count.

A few models deviate from this overall pattern. Notably, \texttt{llama3.3:72b} exhibits a relatively high number of invalid outputs despite its large size, and \texttt{gpt-oss:120b}, which would be expected to perform among the best, also shows unexpectedly poor reliability.
Interestingly, \texttt{llama3.1:8b}, one of the smallest models, has a number of incorrectly generated samples similar compared with huge \texttt{deepseek-r1} model.

Given that the dataset contains only 186 samples and that some models produce up to 116 incorrect outputs, the overall error rates remain substantial. Consequently, future work should explore methods that enforce the generation of semantically and structurally valid evidence. Approaches such as constrained decoding or structured output generation could substantially reduce the occurrence of erroneous outputs~\cite{geng2023grammar}.

\end{description}

\begin{description}[style=unboxed,leftmargin=0em,listparindent=\parindent]
    \setlength\parskip{0em}
    \item[LLMs Extraction Performance.]

\begin{figure}[!t]
    \centering
    \includegraphics[width=\linewidth]{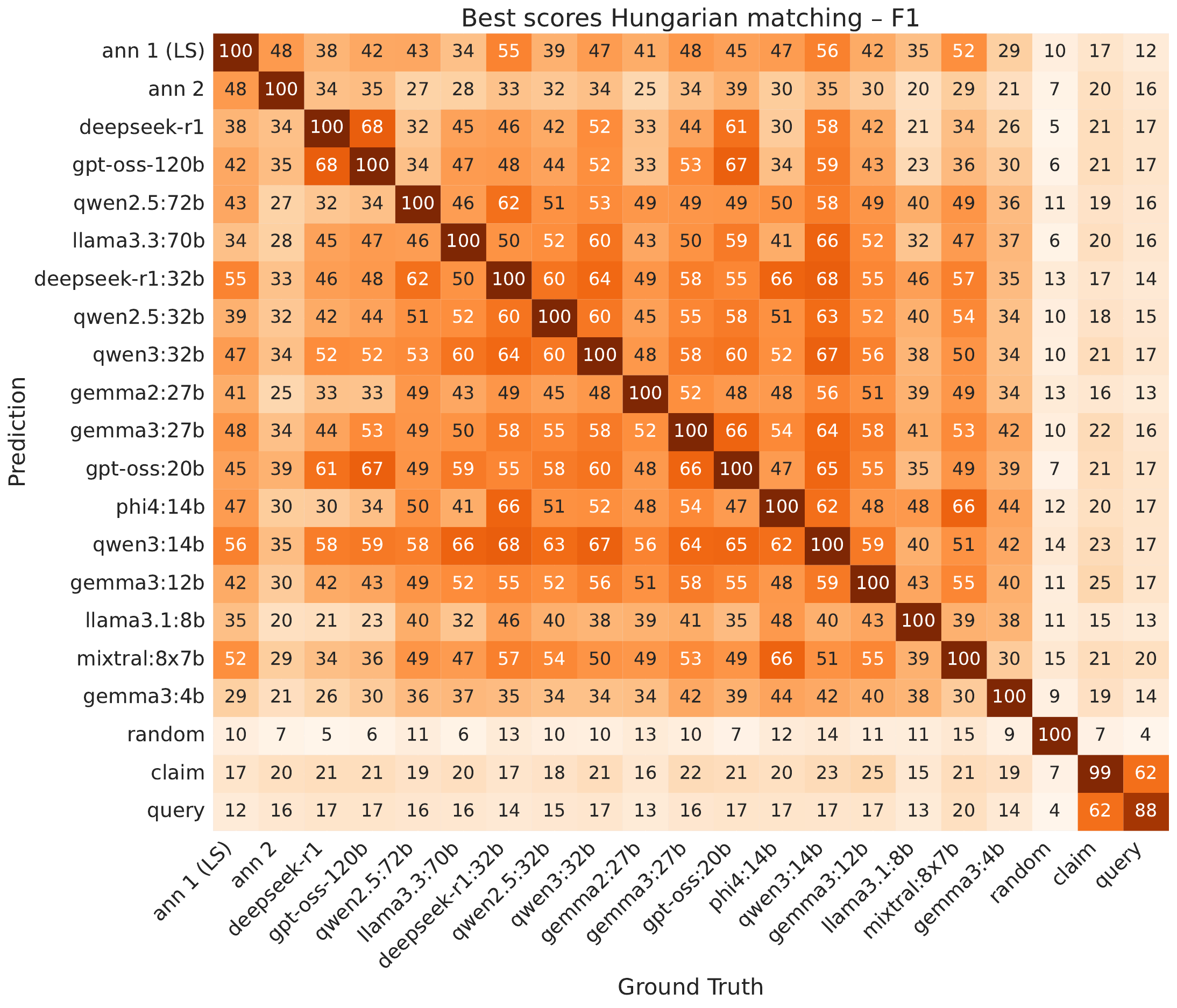}
    \caption{Token-level F1 scores (on a 1–100 scale) between human annotations, large language models (LLMs), and non-neural baseline models. Baselines \texttt{random}, \texttt{claim}, and \texttt{query} are included for comparison. \texttt{ann~1~(LS)} refers to annotations created in Label Studio, while \texttt{ann~2} denotes annotations collected through the custom annotation interface.}
    \label{fig:f1_hung}
\end{figure}

Before presenting the results, it should be emphasized, that certain LLM pairs matches are evaluated using a substantially reduced subset of data.
This is because, as previously discussed, LLMs can fail to generate valid annotations. 
Excluding these missing data points may introduce bias --- for instance, by disproportionately removing more challenging samples and thereby artificially inflating the performance metrics of certain models. Nevertheless, this exclusion is necessary to ensure that model pairs are evaluated on the same set of available samples.

To reduce noise and improve comparability in the evaluation, we remove stop words (see Appendix~\ref{app: stop words} for entire list of stop words) from all evidence sets before computing F1-scores.
The degree of overlap between two annotation sets, is then computed using token-level F1 score.
Since neither human annotators nor LLMs are instructed to produce exhaustive span selections, the number of annotated spans may differ between the two sets.
To not penalize different degree of exhaustiveness, we use the Hungarian matching algorithm~\cite{kuhn1955hungarian} to find optimal assignment between two annotation sets.
First, we compute the F1-score for all possible span pairs across the two sets.
Second, we apply the Hungarian algorithm to solve the assignment problem, ensuring that each span is matched to at most one span in the other set while maximizing the total F1.
Finally, the average F1 of the resulting optimal matching represents the token-level F1 for a single data point.

The average F1-score between two annotators is 48, as we can see in the Figure~\ref{fig:f1_hung}. 
LLMs \texttt{deepseek-r1:32b} and \texttt{qwen3:14b} achieve the strongest performance among the evaluated models, with alignment scores of 55 and 56, respectively, against the annotations created in Label Studio --- surpassing even the human–human agreement. 
However, it should be emphasized that these scores are calculated only on samples where LLM annotations were generated correctly and human annotations were available.

All the non-neural baseline methods proves to be weak, as all of them have F1-score less then 18. 
Further analysis shows, that the precision for the \texttt{claim} and \texttt{query} baselines is high --- around 30, showcasing that words in the claim and query are often also used as evidence by annotators.
However, the recall is very low, therefore decreasing the overall F1 performance significantly. 

Lastly, a comparison between the two sets of human annotations reveals systematic differences. All evaluated LLMs achieve higher alignment with the annotations created in Label Studio, than with those produced in the custom annotation environment, suggesting that the two annotation settings yield slightly different annotation styles or levels of granularity. Further analysis is required to reveal the differences between annotations.

\end{description}

\begin{description}[style=unboxed,leftmargin=0em,listparindent=\parindent]
    \setlength\parskip{0em}
    \item[Alignment with Human Annotators.]
    
\begin{figure}
    \centering
    \includegraphics[width=0.9\linewidth]{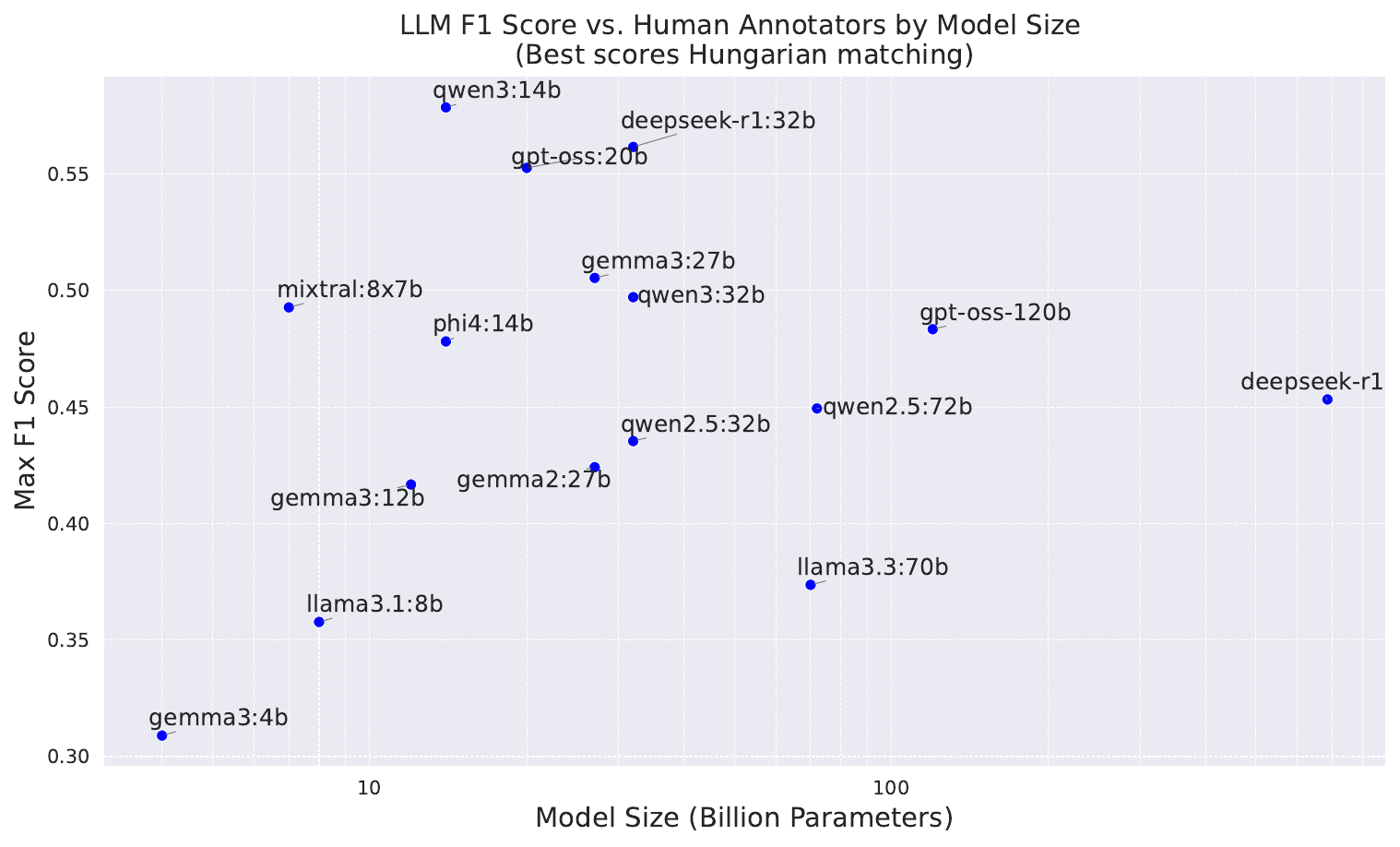}
    \caption{Token-level F1 scores between large language models (LLMs) and human annotations against the model size in billions (log scale). For each data point, the maximum F1 score across the two annotators is used, reflecting alignment with at least one human annotator. }
    \label{fig:match_ann_f1}
\end{figure}

To directly evaluate alignment with human annotators, each LLM-generated evidence span is compared separately with both annotators’ labels. The higher of the two token-level F1 scores is then selected, reflecting agreement with at least one annotator.

The results of this analysis are presented in Figure~\ref{fig:match_ann_f1}, where the maximum token-level F1 score is plotted against model size. 
While performance generally improves from small to medium-sized models, adding substantially more parameters (e.g., 685B \texttt{deepseek-r1} or 120B \texttt{gpt-oss}) does not yield further gains. This suggests that beyond a certain threshold, model size alone does not correlate with better extraction performance.
Nevertheless, the models 14B \texttt{qwen3}, 32B \texttt{deepseek-r1}, and 20B \texttt{gpt-oss} exhibit a favourable trade-off between parameter count and alignment with human annotations.

\end{description}

\section{Conclusion}

We introduce a new dataset of Czech and Slovak texts with fine-grained evidence annotations, produced by two independent annotators for each sample. This dataset enables the computation of inter-annotator agreement, which we measure using the Hungarian matching algorithm and the Token-F1 metric, resulting in a score of 47.
Additionally, it provides a foundation for evaluating the alignment of LLM-generated evidence with human judgments --- filling an existing gap in available resources.

Using this dataset, we analysed the ability of various LLMs to generate valid fine-grained evidence. We observed a clear relationship between model size and the proportion of valid outputs: smaller models such as 4 B \texttt{gemma3} and 8 B \texttt{mixtral} exhibited error rates exceeding 50\%.
This highlights the requirement to employ constrained decoding mechanisms in the further work.

Despite these generation errors, our results on LLMs extraction performance show diminishing returns with increasing model size. Performance improves from small to medium-sized models, but beyond a certain threshold additional parameters do not yield better extraction accuracy. Notably, 14 B \texttt{qwen3}, 32 B \texttt{deepseek-r1} and 20 B \texttt{gpt-oss} offer the best trade-off between model size and alignment with human annotations.

\begin{description}[style=unboxed,leftmargin=0em,listparindent=\parindent]
    \setlength\parskip{0em}
    \item[Acknowledgements.]
    This work was supported by the Technology Agency of the Czech Republic (TAČR) under the SIGMA Programme, 8th Public Competition, Sub-objective 4: Bilateral Cooperation, project TQ16000028.

\end{description}

%
%
\bibliographystyle{splncs04}
\bibliography{bib}

\newpage

\appendix

\section{List of Stop-Words}
\label{app: stop words}

\begin{tabular}{l l l l l l l l l l}
\toprule
aby & dalsi & kam & ne & prvni & tohle & a & když & ní & takže \\
aj & design & kde & nebo & pta & toho & aniž & která & nové & te \\
ale & dnes & kdo & nejsou & re & tohoto & ano & které & nový & tě \\
ani & do & kdyz & neni & si & tom & až & který & o & těma \\
asi & email & ke & nez & strana & tomto & budeš & kteři & ode & této \\
az & ho & ktera & nic & sve & tomuto & být & ku & on & tím \\
bez & jak & ktere & nove & svych & tu & což & máte & práve & tímto \\
bude & jako & kteri & novy & svym & tuto & či & me & proč & toto \\
budem & je & kterou & od & svymi & ty & článek & mě & protože & tvůj \\
budes & jeho & ktery & pak & ta & tyto & článku & mít & první & u \\
by & jej & ma & po & tak & uz & články & mně & před & už \\
byl & jeji & mate & pod & take & vam & další & mnou & přede & v \\
byla & jejich & mezi & podle & takze & vas & i & můj & přes & vám \\
byli & jen & mi & pokud & tato & vase & já & může & při & váš \\
bylo & jeste & mit & pouze & tedy & ve & její & my & s & vaše \\
byt & ji & muj & prave & tema & vice & jenž & ná & se & více \\
ci & jine & muze & pred & ten & vsak & ještě & nám & sice & však \\
clanek & jiz & na & pres & tento & za & jiné & napište & své & všechen \\
clanku & jsem & nad & pri & teto & zda & již & náš & svůj & vy \\
clanky & jses & nam & pro & tim & zde & jseš & naši & svých & z \\
co & jsme & napiste & proc & timto & ze & jšte & nechť & svým & zpět \\
coz & jsou & nas & proto & tipy & zpet & k & není & svými & zprávy \\
cz & jste & nasi & protoze & to & zpravy & každý & než & také & že \\
\bottomrule
\end{tabular}

\newpage
\section{Prompt Used to Extract Fine-Grained Evidence}
\label{app:LLM Prompt}

\begin{figure}[!ht]

    \begin{tcolorbox}[colback=gray!10, colframe=black!70, title=Fine-Grained Evidence Extraction Instructions Prompt]
    \textbf{Comment (for context):} \{\{source\_comment\}\} \\
    \textbf{Claim (extracted from the comment):} \{\{claim\}\} \\
    \textbf{Text:} \{\{text\}\}
    
    \vspace{0.6em}
    Your task is to identify the smallest possible parts of the Text that directly support the claim. 
    
    \vspace{0.3em}
    Focus on the phrases that most clearly justify or confirm the truth of the claim. 
    Avoid selecting entire sentences unless absolutely necessary—choose the shortest meaningful spans that stand on their own. 
    You may select multiple spans if more than one part of the text provides evidence.
    
    \vspace{0.3em}
    Do not modify, correct, or rewrite the text. 
    Preserve all grammatical and syntactic errors exactly as they appear in the original.
    
    \vspace{0.3em}
    Return only a \texttt{JSON} object of type \texttt{Dict[str, List[str]]} 
    with the key \texttt{'spans'} and a list of selected spans as its value.
    
    \vspace{0.3em}
    \textbf{Important:} Every selected span must be an exact substring of the given Text --- character-for-character identical. 
    Do not paraphrase, retype, or alter any character (including punctuation, spacing, or capitalization).
    \end{tcolorbox}
    
    \caption{Prompt used for large language models to extract fine-grained evidence supporting the claim. During the generation, \{\{source\_comment\}\}, \{\{claim\}\} and \{\{text\}\} placeholders were substituted with the source comment containing claim, text of the claim and the text to extract fine-grained relevance from, respectively.  }
    \label{fig:LLM-prompt-fge-supervision}
\end{figure}

\end{document}